\pdfoutput=1
\documentclass[letterpaper, 10 pt, conference]{ieeeconf}  %

\IEEEoverridecommandlockouts                              %

\usepackage[T1]{fontenc}

\usepackage{amsfonts}	%
\usepackage{amsmath}	%
\usepackage{amssymb}    %
\usepackage{siunitx}
\usepackage{xfrac}    %
\usepackage{pifont}   %

\usepackage{booktabs}
\usepackage{makecell}  %
\usepackage[flushleft]{threeparttable}  %
\usepackage{multirow}
\usepackage{rotating}

\usepackage{xspace}    %
\usepackage[usenames,dvipsnames]{xcolor}    %
\usepackage{colortbl}
\let\labelindent\relax
\usepackage[inline]{enumitem} %
\usepackage{graphicx} %
\usepackage{microtype}
\usepackage{cite}
\usepackage{flushend}
\usepackage{xcolor}
\usepackage{svg}
\usepackage{adjustbox}
\usepackage{makecell} %

\newcommand{\cmarkblue}{{\color{green}\ding{51}}}   %

\newcommand{\xmarkred}{{\color{red}\ding{55}}}      %

\makeatletter
\let\NAT@parse\undefined
\makeatother

\usepackage{url}
\usepackage{eso-pic}  %

\usepackage[pdfencoding=auto, colorlinks=true]{hyperref} %
\usepackage[hang,flushmargin]{footmisc}

\usepackage[capitalize]{cleveref}
\crefname{section}{Sec.}{Secs.}
\Crefname{section}{Section}{Sections}
\Crefname{table}{Table}{Tables}
\crefname{table}{Tab.}{Tabs.}

\begin{document}

\newcommand{\myfunding}{This work has been supported by projects PID2021-126623OB-I00 and PID2024-161576OB-I00, funded by MCIN/AEI/10.13039/501100011033 and co-funded by the European Regional Development Fund (ERDF, “A way of making Europe”), by project PLEC2023-010343 (INARTRANS 4.0) funded by MCIN/AEI/10.13039/501100011033, and by the R\&D program TEC-2024/TEC-62 (iRoboCity2030-CM) and ELLIS Unit Madrid, granted by the Community of Madrid.}

\title{\LARGE \bf
GaussianCaR: Gaussian Splatting for Efficient Camera-Radar Fusion
}

\author{
Santiago Montiel-Marín$^{1}$,
Miguel Antunes-García$^{1}$,
Fabio Sánchez-García$^{1}$, \\
Angel Llamazares$^{1}$,
Holger Caesar$^{2}$,
and Luis M. Bergasa$^{1}$%
\thanks{$^{1}$ Department of Electronics. University of Alcalá, Spain.}%
\thanks{$^{2}$ Department of Cognitive Robotics. Delft University of Technology, The Netherlands.}%
\thanks{\myfunding}%
}

\maketitle

\AddToShipoutPictureBG*{%
  \AtPageLowerLeft{%
    \raisebox{0.28in}{\makebox[\paperwidth]{%
      \parbox{\textwidth}{\centering\scriptsize
        Accepted to the 2026 IEEE International Conference on Robotics and Automation (ICRA).
        Published version: \href{https://ieeexplore.ieee.org/document/11697071}{IEEE Xplore},
        DOI: \href{https://doi.org/10.1109/ICRA57385.2026.11697071}{10.1109/ICRA57385.2026.11697071}.\\
        \copyright~2026 IEEE. Personal use of this material is permitted. Permission from IEEE must be
        obtained for all other uses.}}}}}
\thispagestyle{empty}
\pagestyle{empty}

\begin{abstract}
    Robust and accurate perception of dynamic objects and map elements is crucial for autonomous vehicles performing safe navigation in complex traffic scenarios. While vision-only methods have become the de facto standard due to their technical advances, they can benefit from effective and cost-efficient fusion with radar measurements. In this work, we advance fusion methods by repurposing Gaussian Splatting as an efficient \textit{universal view transformer} that bridges the view disparity gap, mapping both image pixels and radar points into a common Bird's-Eye View (BEV) representation. Our main contribution is GaussianCaR, an end-to-end network for BEV segmentation that, unlike prior BEV fusion methods, leverages Gaussian Splatting to map raw sensor information into latent features for efficient camera-radar fusion. Our architecture combines multi-scale fusion with a transformer decoder to efficiently extract BEV features. Experimental results demonstrate that our approach achieves performance on par with, or even surpassing, the state-of-the-art on BEV segmentation tasks (57.3\%, 82.9\%, 50.1\% IoU for vehicles, roads, and lane dividers) on the nuScenes dataset, while maintaining a 3.2× faster inference runtime. \href{https://www.github.com/santimontiel/gaussiancar}{Code} and \href{https://www.santimontiel.eu/projects/gaussiancar}{project page} are available online.
\end{abstract}

\section{Introduction}
\label{sec:intro}

Developing robust and accurate perception models within an efficient framework is a cornerstone to enabling autonomous vehicles to achieve reliable scene understanding. Effectively interpreting dynamic objects and static map elements is a step towards ensuring safe navigation in complex environments such as traffic scenarios. Early perception solutions relied on LiDAR measurements \cite{lang2019pointpillars,yin2021center} due to the high precision of their 3D geometric information, despite the high cost and sensitivity to adverse weather conditions. With the advent of Deep Learning-based \textit{projectors} or \textit{view transformation} modules, which map image pixels into 3D space, vision-centric solutions \cite{philion2020lss,huang2021bevdet,li2022bevformer,harley2023simplebev} emerged as the dominant paradigm, offering a cost-effective path to large-scale deployment of perception models in the automotive domain. However, while cameras provide rich and dense semantic information, they lack motion cues and precise geometric accuracy, leading to depth and scale ambiguities, as well as localization errors. In contrast, radar provides a sparse yet accurate point cloud with position and velocity measurements, making it a suitable complementary sensor to cameras. Fusing cameras and radar signals enables a robust and cost-effective perception framework, suitable for mass-scale deployment in autonomous systems.

\begin{figure}[t]
  \centering
  \includegraphics[width=0.95\linewidth]{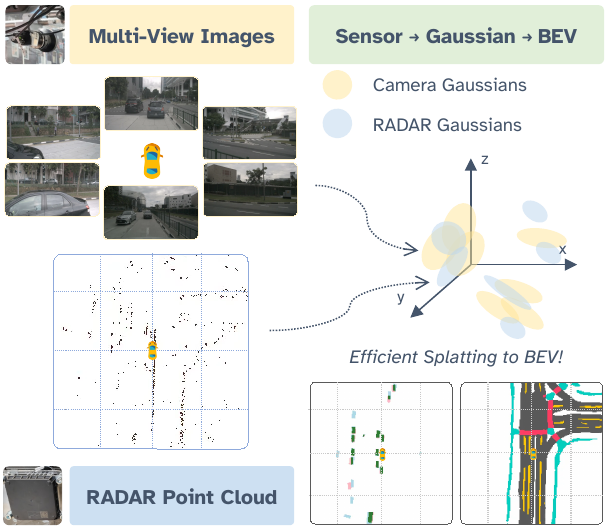}
  \caption{We propose \textbf{GaussianCaR}, a novel method for efficient camera-radar fusion. We envision sensor fusion as a \textbf{modality} $\rightarrow$ \textbf{Gaussians} $\rightarrow$ \textbf{BEV} transformation, achieving competitive accuracy with significantly fast inference times for BEV segmentation tasks.}
  \label{fig:motivation}
  \vspace{-0.3cm}
\end{figure}

In this work, we tackle the key problem of fusing camera and radar modalities
to produce a dense and robust Bird's-Eye View (BEV) latent representation within a simple yet
efficient framework for BEV perception tasks, focusing on vehicle
and map segmentation. Since the introduction of BEVFusion \cite{liu2023bevfusion},
sensor fusion through BEV latent representations has become standard practice.
The main challenge to fuse multiple modalities with different input
representations lies in bridging the \textit{view disparity}. On the one hand,
camera data is represented as images, which naturally lack depth, scale, and
motion information. To mitigate this issue, recent literature identifies two
trends for performing an image-to-BEV transformation. Depth or forward-based
approaches \cite{philion2020lss,huang2021bevdet,li2023bevdepth} estimate a
depth distribution along rays passing through each pixel, but feature projection
is limited to the physical distribution of grid cells. Projection or
backward-based transformations aim to pull image features to a volume through
simple interpolation \cite{harley2023simplebev} or costly attention-based learning
\cite{li2022bevformer}. On the other hand, radar point clouds must also be
transformed into BEV representations, mainly through voxelization or pillarization.
Mapping a radar measurement to a grid cell is straightforward; however, each point
is assigned to a single voxel or pillar, without taking into account the uncertainties
in the measurements. This process results in highly sparse latent BEV representations
due to the limited number of points in radar point clouds and the relatively large
working grid size. Recently, 3D Gaussian Splatting (GS) \cite{kerbl20233d} has emerged
as an efficient and powerful technique for 3D scene reconstruction, representing a
scene as a set of learnable Gaussians that can be differentiably rasterized into
plane-like representations. Inspired by this, we envision BEV sensor fusion as a
\textbf{modality/view $\rightarrow$ Gaussians $\rightarrow$ BEV} transformation,
leveraging GS as a \textit{universal view transformer} for all modalities. This approach enables the unified sensor fusion of diverse inputs (pixels and points)
with dense feature propagation and uncertainty awareness.

The main contribution of this paper is a robust, simple, and efficient sensor
fusion framework for camera and radar data, leveraging GS
as a universal view transformer. We propose \textbf{GaussianCaR}, a novel
method for BEV segmentation that uses two modality-specific encoders --
Pixels-to-Gaussians and Points-to-Gaussians -- to lift features from each sensor
space into a unified sparse 3D space, enabling multi-modal fusion. To the best of our knowledge, we are the first model dedicated to BEV segmentation fusing camera and radar data within a Gaussian-based framework. Finally, we perform a multi-stage transformer-based fusion and decoding process to produce the desired BEV outputs. Extensive evaluation on the nuScenes dataset \cite{caesar2020nuscenes} demonstrates that our approach achieves state-of-the-art (SOTA) performance on dense BEV perception tasks while maintaining efficient inference time and memory usage.

In summary, we make three key claims: (i) \textbf{GaussianCaR} scores on par with, or even surpasses, SOTA methods in dense BEV perception tasks, such as vehicle, drivable surface, and lane segmentation; (ii) our Pixels-to-Gaussians and Points-to-Gaussians modules efficiently lift modality features to BEV space, enabling effective multi-modal sensor fusion; and (iii) the method is fast and efficient in terms of inference time, making it suitable for deployment. These claims are supported by the results and evaluations presented in this manuscript.

\section{Related Work}
\label{sec:related}

In this section, we review related works in three areas: camera-based BEV perception, camera-radar fusion for BEV perception, and the use of GS in robotics.

{\parskip=3pt
\noindent\textbf{Camera-based BEV Perception.} Vision-centric solutions for perception tasks were fundamentally limited by the ill-posed nature of monocular depth estimation in the camera perspective view. The foundational work LSS \cite{philion2020lss} proposed a shift from perspective view to local camera frustum space via differentiable \textit{feature lifting} by predicting depth distribution and features per image. Lifted features from multiple views are aggregated into a unified BEV space. BEVDepth \cite{li2023bevdepth} improved depth estimation by incorporating cross-modal supervision from sparse LiDAR measurements. The BEVDet series \cite{huang2021bevdet,huang2022bevpoolv2} further enhanced performance and introduced efficient view transformation techniques.

Other methods rely on projection or learning-based approaches performing view transformation via a learned component or an attention mechanism. SimpleBEV \cite{harley2023simplebev} employs bilinear sampling to populate local camera frustums. BEVFormer \cite{li2022bevformer} introduces a 2D-to-3D attention mechanism to associate a set of BEV queries with image features. Hybrid approaches, such as FB-Occ \cite{li2023fb} and BEVNeXt \cite{li2024bevnext}, combine geometrical and learning-based transformations within a unified framework.

In this work, we implement a Pixels-to-Gaussians encoder that lifts camera features to BEV space via differentiable Gaussian rasterization, expanding geometrical-based view transformations with a coarse-to-fine strategy for accurate spatial positioning of Gaussians in metric space.
}

{\parskip=3pt
\noindent\textbf{Camera-Radar Fusion for BEV Perception.} Camera and radar sensors exhibit complementary strengths and weaknesses. Cameras provide dense, high-resolution semantic information, while radar delivers sparse but reliable spatial and motion cues, especially under adverse conditions. Fusing both modalities enables more accurate and robust scene understanding.

Early fusion methods operated in the perspective view by projecting points onto the image plane, as seen in CenterFusion \cite{nabati2021centerfusion}, RADIANT \cite{long2023radiant}, and CRAFT \cite{kim2023craft}. With the emergence of BEVFusion \cite{liu2023bevfusion}, dense fusion in a unified BEV space became the dominant paradigm for combining images and radar point clouds. SimpleBEV \cite{harley2023simplebev} incorporated radar in BEV space via an occupancy map. BEVCar \cite{schramm2024bevcar} uses radar points to guide bilinear sampling of image features, enabling multi-level fusion. Methods such as CRN \cite{kim2023crn} and CRT-Fusion \cite{kim2024crt} perform fusion in the camera frustum view using fused frustum volumes or attention-based mechanisms to align modalities.

We propose performing fusion in BEV space through a two-step transformation, \textbf{modality} $\rightarrow$ \textbf{Gaussians} $\rightarrow$ \textbf{BEV}. To this end, we encode radar data with our Points-to-Gaussians module, which converts radar measurements to Gaussians to effectively capture and propagate uncertainty.

}

\begin{figure*}
    \centering
    \vspace{2mm}
    \includegraphics[width=1.0\textwidth]{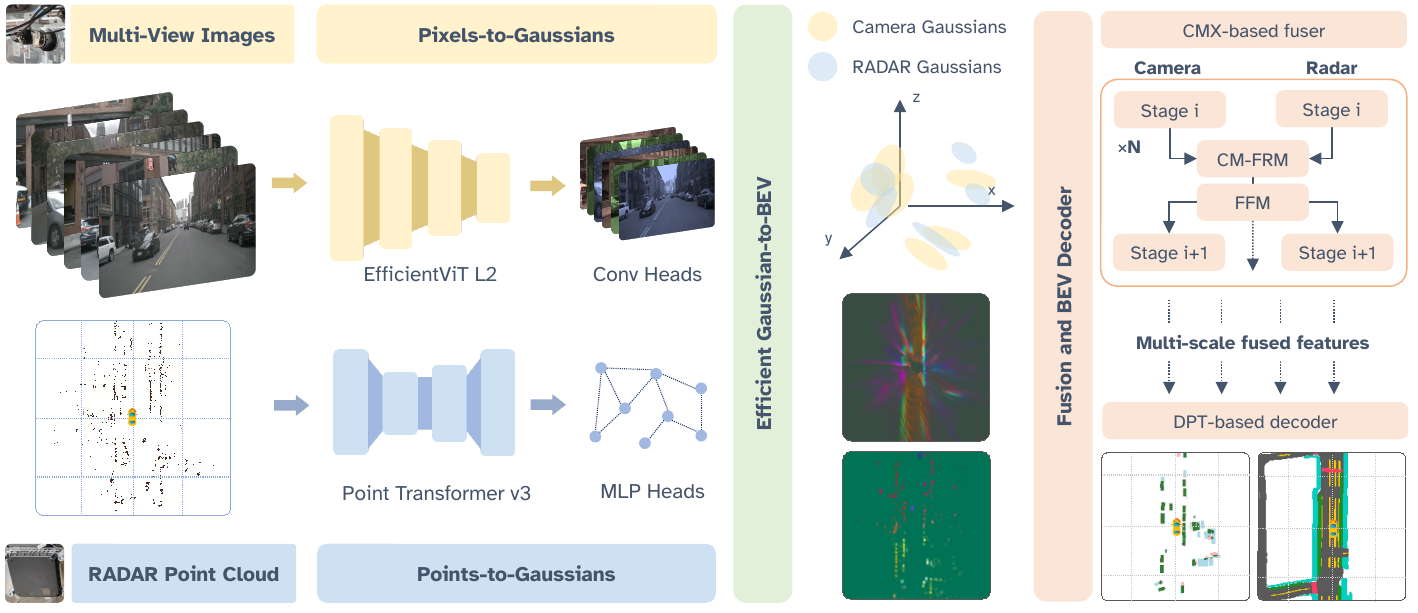}
    \caption{
        \textbf{Main diagram of our proposal, GaussianCaR.} Given
        \textcolor{YellowOrange!70}{multi-view camera images} and
        \textcolor{NavyBlue}{radar point clouds}, we leverage GS as a
        \textit{universal view transformer} and formulate sensor fusion as
        \textbf{modality} $\rightarrow$ \textbf{Gaussians} $\rightarrow$ \textbf{BEV} transformation. The model
        predicts \textcolor{BurntOrange}{BEV segmentation maps} for dynamic vehicles
        and map elements. We employ two feature encoding branches:
        \textcolor{YellowOrange!70}{Pixels-to-Gaussians} for camera features and \textcolor{NavyBlue}{Points-to-Gaussians} for radar
        point clouds. Features are splatted and fused in BEV space using a \textcolor{BurntOrange}{CMX-based}
        fuser, and decoded via a \textcolor{BurntOrange}{DPT} decoder.
    }
    \label{fig:architecture}
\end{figure*}

{\parskip=3pt
\noindent\textbf{Gaussian Splatting in Robotics.} GS \cite{kerbl20233d} has rapidly become a foundation for scene reconstruction and neural rendering. It represents 3D environments as sets of learnable anisotropic Gaussian primitives, each parameterized by position, scale, rotation, opacity, and feature vector. This formulation enables efficient, fully differentiable forward rendering and continuous geometric representation.

The ability to capture the environment with an efficient, continuous, and differentiable geometric representation makes GS a promising approach for robotics and perception applications. In the SLAM domain, OpenGS-SLAM \cite{yang2025opengs} performs open-set segmentation and indoor scene reconstruction from an \mbox{RGB-D} stream as input, while WildGS-SLAM \cite{zheng2025wildgs} reconstructs 3D Gaussian maps for static scenes, handling dynamic objects to avoid scene blurring. In dense BEV perception, GaussianLSS \cite{lu2025toward} and GaussianBeV \cite{chabot2025gaussianbev} introduce camera-only architectures with differentiable Gaussian rendering to lift image features into BEV space and perform semantic segmentation in an end-to-end fashion.

Building upon this paradigm, we repurpose GS as a \textit{universal view transformer}, mapping input modalities to BEV latent representations through a set of 3D Gaussian primitives, and extend the approach from camera to camera-radar data. To the best of our knowledge, this is the first cost-effective and efficient Gaussian-based framework for camera-radar sensor fusion applied to BEV segmentation tasks, paving the way for large-scale deployment.
}

\section{Methodology}
\label{sec:main}

\subsection{Task Definition and Overview}

The primary objective of \textbf{GaussianCaR}, described in Fig. \ref{fig:architecture}, is to predict BEV segmentation maps of relevant road entities for autonomous navigation, such as vehicles or drivable surfaces, leveraging GS techniques to fuse multi-view cameras and radar sensors.

Given as inputs: (a) images from a multi-camera system with $N_c$ views, $I\:\in\:\mathbb{R} ^ {N_c\times 3 \times H \times W}$; (b) a radar point cloud with $N_r$ points, and $F_r$ dimensional features, $R\in\mathbb{R}^{N_r\times F_r}$; and (c) the intrinsic and extrinsic calibration matrices between the vehicle sensors. From these inputs, GaussianCaR produces a BEV segmentation map, $S \in \mathbb{R}^{C \times H_{BEV} \times W_{BEV}}$, where $C$ is the number of semantic classes and $H_{BEV}$, $W_{BEV}$ define the BEV resolution. Our approach leverages two modality-specific encoders: for camera, Pixels-to-Gaussians (Sec. \ref{subsec:pixel}), and for radar, Points-to-Gaussians (Sec. \ref{subsec:point}). Our modality-based fusion and BEV decoding is described in Sec. \ref{subsec:fusion}. Our training objectives are defined in Sec. \ref{subsec:losses}.

\begin{figure*}[t]
    \centering
    \vspace{2mm}
    \includegraphics[width=1.00\textwidth]{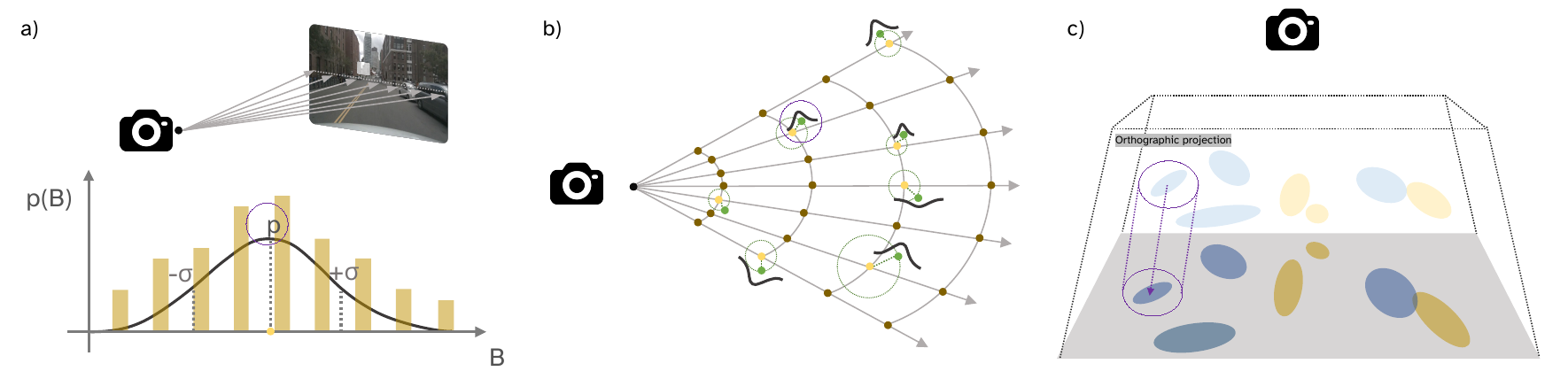}
    \caption{
        \textbf{Gaussian modeling process.} In (a), we present the process of extracting a Gaussian from a discrete probability distribution; in (b), we depict the behavior of the offset head, displacing the final Gaussian position from the original set of candidates; in (c), we illustrate the Gaussian rasterization process, projecting Gaussians from 3D space to BEV space via orthographic projection.
    }
    \label{fig:extra}
\end{figure*}

\subsection{Pixels to Gaussians} \label{subsec:pixel}

\begin{figure}[t]
  \centering
  \includegraphics[width=1.0\linewidth]{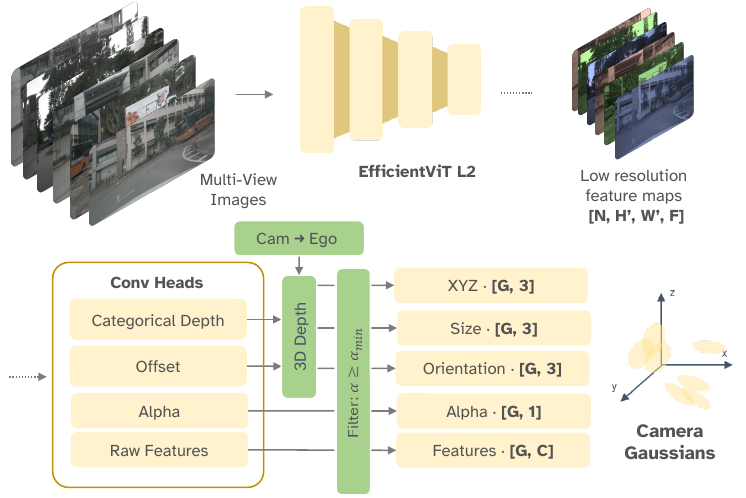}
  \caption{Our \textbf{Pixels-to-Gaussians} extracts low-resolution feature maps using an EfficientViT backbone and a neck. A set of convolutional heads predicts $\mathcal{G}_c$ Gaussians. To position the Gaussians in 3D space, camera intrinsic and extrinsic matrices are used.}
  \label{fig:pixels}
  \vspace{-0.3cm}
\end{figure}

For our Pixels-to-Gaussians encoder (in Fig. \ref{fig:pixels}), we build upon the foundations in \cite{chabot2025gaussianbev,lu2025toward}. We process multi-camera images, $I$, and extract x1/8 low-resolution feature maps, $F$, using EfficientViT-L2 \cite{cai2023efficientvit}, a lightweight, transformer-based backbone, and a neck for feature aggregation.

To lift image features from pixel space to 3D, we map from pixels to Gaussians, producing $\left| \mathcal{G} \right| = N_c\cdot H_{low}\cdot  W_{low}$ Gaussians. A series of convolutional heads is applied to the low-resolution feature maps to predict the physical and semantic properties of each Gaussian, $\mathcal{G}_i$, including position, $p_i$, size, $s_i$, orientation, $R_i$, opacity, $\alpha_i$, and features, $f_i$.

We estimate the geometrical position or mean of each Gaussian, $p_i$, using a coarse-to-fine strategy. In the coarse stage, a depth head predicts a probability distribution along the optical ray for each pixel, where depth is uniformly discretized into $B$ bins between $\left[ d_{min}, d_{max} \right]$. This produces a tensor $F_{dep}\in \mathbb{R}^{\left| \mathcal{G} \right| \times B}$, containing per-Gaussian depth classification logits. A coarse position is calculated via probability-weighted sum over the depth bins and projected to 3D space using the camera intrinsic and extrinsic matrices. In the fine stage, an offset head refines the final 3D position in metric space, $F_{off} \in \mathbb{R}^{\left| \mathcal{G} \right| \times 3}$, enabling the Gaussian to deviate from the set of discrete bin centers and achieve higher precision. Final per-Gaussian position (shown in Fig. \ref{fig:extra}.a-b) is determined as:
\begin{equation}
\mathbf{p}_i = \mathcal{P}(\mathbf{u}_i, \hat{d}_i(F_{dep_i})) + F_{off_i}
\end{equation}

\noindent where $\hat{d}_i$ is the predicted bin center along the optical ray, and $\mathcal{P}(\mathbf{u}_i, \dots)$ is the back-projection of a pixel $\mathbf{u}_i$ to the world using intrinsic and extrinsic matrices.

The size, $s_i$, and orientation, $R_i$, of each Gaussian are derived from the predicted depth distribution and camera geometry. From the probability distribution in the coarse estimation, we compute the standard deviations around the mean position. These deviations are assembled into a covariance matrix that encodes spatial uncertainty. The covariance is then scaled by an error tolerance coefficient, $k=0.5$, which controls the effective spread of the Gaussian. Finally, the eigenvalues of the scaled covariance determine the size along each principal axis, while the eigenvectors define the orientation in 3D space.

Lastly, each Gaussian is assigned with an opacity parameter $\alpha_i \in \left[ 0, 1 \right]$, predicted by a convolutional head followed by a sigmoid activation, yielding a tensor $F_{opa} \in \mathbb{R}^{\left| \mathcal{G} \right| \times 1}$. This parameter regulates the influence of each Gaussian during differentiable rendering. Furthermore, we empirically set a minimum threshold $\alpha_{min} = 0.01$, allowing Gaussians with negligible contribution to be discarded.

\subsection{Points to Gaussians} \label{subsec:point}

\begin{figure}[t]
  \centering
  \includegraphics[width=1.0\linewidth]{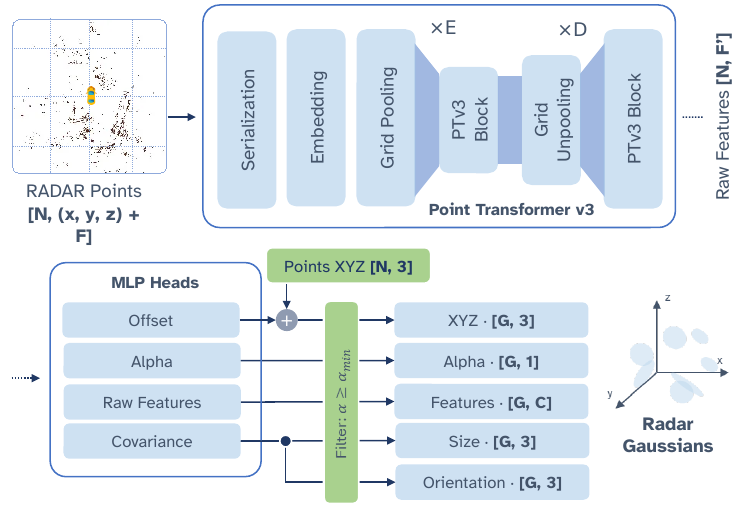}
  \caption{Our proposed \textbf{Points-to-Gaussians} module processes radar point
clouds using a lightweight PTv3, composed of $\mathcal{E}$
encoder and $\mathcal{D}$ decoder blocks. A set of MLP heads then predicts
$\mathcal{G}_r$ Gaussians, each parameterized by geometric and semantic
attributes.}
\label{fig:points}
  \vspace{-0.3cm}
\end{figure}

To extract features from radar point clouds, we employ a lightweight variant of Point Transformer v3 (PTv3) \cite{wu2024point}, depicted in Fig. \ref{fig:points}. The raw, unstructured point cloud is serialized and transformed into multiple ordered representations using space-filling curves and neighbor mapping. From these representations, non-overlapping patches are constructed to capture local neighborhoods. An efficient inter-patch attention mechanism is then applied, enabling both spatial and global context modeling at the per-point level. The overall architecture follows a UNet-like design and outputs point-wise feature embeddings with rich semantic information.

On top of these embeddings, we attach a set of MLP heads to predict the physical attributes and semantic properties of each point. Similar to the coarse-to-fine position estimation in the camera branch, we predict only a metric offset head, as the initial point positions are already known. The opacity attribute is estimated in the same manner as in the camera branch. For size and orientation, we predict a compact representation of the covariance matrix:
\begin{equation*}
    R_{cov_{i}} \in \mathbb{R}^{6} = [xx\; xy\; xz\; yy\; yz\; zz]
\end{equation*}

\noindent where eigenvalues are enforced to be positive via a softplus activation. Further implementation details for PTv3 can be found in \cite{wolters2024sparc}.

\subsection{Modality-based Fusion and BEV Decoding} \label{subsec:fusion}
To complete the \textbf{modality $\xrightarrow[]{}$ Gaussian $\xrightarrow[]{}$ BEV} cycle, we splat each set of learned $|f_{i}|$-dimensional Gaussian representations (with $|f_{i}| = 128$ in our experiments) to BEV using differentiable Gaussian rasterization through an orthographic projection and alpha-blending:
\begin{equation} \label{equa:blending}
\mathbf{F} = \sum_{i \in \mathcal{N}} f_i \, \alpha_i  \prod_{j=1}^{i-1} (1 - \alpha_j)
\end{equation}

\noindent where $f_i$ is the feature vector of each Gaussian, and $F$ is the computed per-pixel feature after blending, shown in Fig. \ref{fig:extra}-c.

Inspired by \cite{zhang2023cmx}, we adopt a four-stage, multi-scale feature fusion strategy, depicted in Fig. \ref{fig:architecture}. Each stage receives feature maps from both modalities and consists of Cross-Modal Feature Rectification and Feature Fusion Modules, producing fused representations that serve as inputs for a DPT-based decoder \cite{ranftl2021vision} to generate the final BEV representation. The output of the first fusion stage is connected to an auxiliary head, while the output of the decoder feeds the final segmentation head.

\begin{table}[t!]
\vspace{2mm}
\centering

\caption{BEV Vehicle Segmentation on the\\nuScenes Validation Set}
\label{tab:sota}
\begin{threeparttable}
\begin{tabular}{@{}lcccccc}
\toprule
\multicolumn{1}{c}{\textbf{Method}} & \textbf{Code}
& \textbf{Cam Enc} & \textbf{Radar Enc} & \textbf{IoU ($\uparrow$)} \\ \midrule
\textit{Camera-only}                  &   &           &          &              \\ \midrule
BEVFormer \cite{li2022bevformer}                  & \cmarkblue & RN-101          & -         & 43.2             \\
GaussianLSS \cite{lu2025toward}                   & \cmarkblue & RN-101          & -         & 46.1             \\
SimpleBEV \cite{harley2023simplebev}              & \cmarkblue & RN-101          & -         & 47.4             \\
PointBeV \cite{chambon2024pointbev}               & \cmarkblue & EN-b4           & -         & 47.8             \\
GaussianBeV \cite{chabot2025gaussianbev}          & \xmarkred  & EN-b4           & -         & 50.3             \\ \midrule
\textit{Camera-radar}                  &   &           &          &              \\ \midrule
SimpleBEV++$^{\ddag}$ \cite{harley2023simplebev}  & \cmarkblue & RN-101          & PFE+Conv  & 52.7             \\
SimpleBEV \cite{harley2023simplebev}              & \cmarkblue & RN-101          & Conv      & 55.7             \\
BEVCar$^{\ddag}$ \cite{schramm2024bevcar}         & \cmarkblue &
\makecell{DINOv2/B \\ +Adapter} & PFE+Conv  & 58.4             \\
CRN$^{\diamond}$ \cite{kim2023crn}                & \xmarkred  & RN-50           & SECOND    & \underline{58.8} \\
BEVGuide \cite{man2023bev}                        & \xmarkred  & EN-b4           & SECOND    & \textbf{59.2}    \\ \midrule \rowcolor{gray!15}
\textbf{GaussianCaR} (\textit{ours})              & \cmarkblue & EViT-L2         & PTv3      & 57.3             \\ \bottomrule
\end{tabular}

\footnotesize{
We report the results from \cite{schramm2024bevcar,chabot2025gaussianbev}. Evaluation done with image resolution $(448, 800)$ (or $^{\ddag}(448, 896)$, if needed) and applying visibility filtering. $^{\diamond}$CRN uses 4 input frames (3 past and 1 current) at inference time. \xmarkred\:GaussianBeV, CRN, and BEVGuide do not release code for BEV segmentation. Best is marked in \textbf{bold} and second best is \underline{underlined}.
}
\end{threeparttable}
\vspace*{-0.4cm}
\end{table}

\subsection{Training Losses} \label{subsec:losses}

We train our model end-to-end using two semantic segmentation loss terms, a main loss $L_{sem}$ and an auxiliary loss $L_{sem}^{aux}$. It is defined as:
\begin{equation}
    L = L_{sem} + L_{sem}^{aux}
\end{equation}

For each component, we adopt a combo loss, comprising a binary cross-entropy $L_{bce}$ and Dice loss $L_{dice}$, and additional centerness $L_{ctr}$ and offset $L_{off}$ components for regularization, each balanced by its respective weight $\lambda_i$.
\begin{align} \label{eq:loss}
    L_{sem} = L_{sem}^{aux} &= \lambda_{bce} \cdot L_{bce}
    + \lambda_{dice} \cdot L_{dice} \notag \\
    &\quad + \lambda_{ctr} \cdot L_{ctr}
    + \lambda_{off} \cdot L_{off}
\end{align}

While $L_{sem}$ is applied to the final BEV prediction and $L_{sem}^{aux}$ is attached to the output of the first feature fusion stage to provide early supervision, both losses are computed using an identical definition, following Eq. \ref{eq:loss}.

\section{Experimental Evaluation}
\label{sec:exp}
The main focus of this work is to enable robust and efficient fusion of camera and radar data for BEV perception tasks, leveraging GS as \textit{universal view transformer}. We present our experiments to demonstrate the capabilities of our method and support our key claims, showing that our approach achieves performance on par with, or even surpassing, SOTA methods in BEV segmentation tasks, while maintaining fast and efficient inference runtimes. Finally, we validate our design choices through an ablation study that highlights the effectiveness of our proposal.

\subsection{Experimental Settings}
\label{ssec:experimental-settings}

We present our experimental setup, detailing the dataset, evaluation metrics, and implementation specifics.

{\parskip=3pt
\noindent\textbf{Dataset and Metrics:}
We train and evaluate our model on the nuScenes~\cite{caesar2020nuscenes} dataset, the only large-scale multi-modal dataset that includes synchronized data from 6 surround-view cameras, 5 automotive radars, and a 32-beam LiDAR, as well as high-quality annotations for 3D objects and map surfaces. The dataset contains 1,000 20-second driving scenes, split into 700 training, 150 validation, and 150 test sequences.

We quantify the performance of our network in the tasks of BEV vehicle and map segmentation using the Intersection over Union (IoU) or Jaccard index metric, defined as:
\begin{equation} \label{eq:iou}
    IoU(\hat{y}, y) = \dfrac{|\hat{y} \cap y|}{|\hat{y} \cup y|} = \dfrac{\sum_{H,W} \hat{y}\cdot y}{\sum_{H,W} (\hat{y} + y - \hat{y} \cdot y)}
\end{equation}
where $\hat{y}_i \in \{0,1\}$ is the confidence-thresholded prediction, and $y_i \in \{0,1\}$ is the ground-truth label.
}

\begin{table}[t]
\vspace{2mm}
\centering
\caption{BEV Map Segmentation on the\\nuScenes Validation Set}
\label{tab:sota-map}
\begin{threeparttable}
\begin{tabular}{@{}lcc}
\toprule
\hspace{1.5cm}\textbf{Method} \hspace{1.5cm}
& \textbf{Driv. Area ($\uparrow$)}
& \textbf{Lane Div. ($\uparrow$)} \\ \midrule
\textit{Camera-only} & & \\ \midrule
LSS \cite{philion2020lss}                         & 72.9             & 20.0             \\
BEVFormer \cite{li2022bevformer}                   & 80.1             & 25.7             \\
GaussianBeV \cite{chabot2025gaussianbev}                     & 82.6             & \underline{47.4} \\ \midrule
\textit{Camera-radar} & & \\ \midrule
BEVGuide \cite{man2023bev}                    & 76.7             & 44.2             \\
SimpleBEV++$^{\ddag}$ \cite{harley2023simplebev}  & 81.2             & 40.4             \\
BEVCar$^{\ddag}$ \cite{schramm2024bevcar}          & \textbf{83.3}    & 45.3             \\ \midrule \rowcolor{gray!15}
\textbf{GaussianCaR} (\textit{ours})               & \underline{82.9} & \textbf{50.1}    \\ \bottomrule
\end{tabular}

\footnotesize{
We report the results from \cite{schramm2024bevcar,chabot2025gaussianbev}.
Evaluation done with image resolution $(448, 800)$ (or $^{\ddag}(448, 896)$, if needed)
and applying visibility filtering. Visibility filtering does not apply to map
evaluation. Best is marked in \textbf{bold} and second best is \underline{underlined}.
}
\end{threeparttable}
\vspace*{-0.4cm}
\end{table}

{\parskip=3pt
\noindent\textbf{Implementation Details:}
We train our model for 40 epochs in a distributed setup consisting of 4x NVIDIA A100 80 GB GPUs, using a DDP strategy and gradient accumulation for an effective batch size of 16. We use the AdamW optimizer and a linear annealing scheduler with warmup. Maximum learning rate is $lr_{max}=3\times10^{-4}$ and linearly decreases to $lr_{end}=0$. Weight decay is set to $w_d=1\times10^{-7}$. GS rasterizers use a version of \texttt{diff-gaussian-rasterization} library from \cite{lu2025toward}. The architecture and codebase are implemented in PyTorch 2.4.1 and Lightning.

Images are processed in half-scale $H, W=(448, 800)$. We apply image data augmentation, such as random horizontal flip, zoom-in/out, and rotations, with camera intrinsic matrices being updated consistently. We accumulate 7 radar sweeps and preprocess all variables in the point cloud. We apply data augmentation in BEV space, following \cite{chambon2024pointbev}.

Gaussians are rasterized to a BEV grid space with a perception range of 100 m in both x-y directions (from -50 to 50 m) with 0.5 m of resolution, resulting in a grid of 200×200 cells.
}

\begin{table}[t]
\vspace{2mm}
\centering
\caption{Ablation Study} \label{tab:ablation}
\begin{threeparttable}
\begin{tabular}{@{}lccc}
\toprule
\multicolumn{1}{c}{\textbf{Method}} & IoU ($\uparrow$) & ms ($\downarrow$) & FPS ($\uparrow$)\\ \midrule
\textbf{Baseline:} GaussianLSS \cite{lu2025toward}            & 46.1 & 53.9 & 18.6 \\ \midrule \rowcolor{yellow!15}
\textit{Image Encoding Branch}                   &      &      &      \\ \midrule
+ EffViT L2                                      & 47.3 & 56.6 & 17.8 \\
+ Offset Head                                    & 47.7 & 56.9 & 17.6 \\
+ Early auxiliary loss                           & 47.8 & 56.9 & 17.6 \\
+ Dice loss                                      & 48.0 & 56.9 & 17.6 \\ \midrule \rowcolor{blue!15}
\textit{Radar Encoding Branch}                   &      &      &      \\ \midrule
+ PTv3 w/ scatter (XYZ)                         & 55.0 & 86.1 & 11.6 \\
+ PTv3 w/ scatter (\textit{all variables})      & 56.1 & 86.9 & 11.5 \\
+ PTv3 w/ Gaussians (\textit{all variables})    & 56.9 & 83.7 & 12.0 \\ \midrule  \rowcolor{orange!15}
\textit{Fusion and BEV Decoding}                 &      &      &      \\ \midrule
+ DPT-based decoder                              & 57.1 & 78.2 & 12.8 \\
+ CMX-based fuser                                & 57.3 & 75.6 & 13.2 \\ \bottomrule
\end{tabular}

\footnotesize{
All experiments are run on an NVIDIA RTX 4090 with an image resolution of $(448, 800)$ for the task of vehicle segmentation, utilizing visibility filtering.
}
\end{threeparttable}
\vspace*{-0.4cm}
\end{table}

\subsection{Quantitative Results}
\label{ssec:quantitative-results}

In this section, we compare \textbf{GaussianCaR} with our camera-only baseline, GaussianLSS \cite{lu2025toward}, and SOTA approaches across two BEV segmentation tasks: vehicle and map. We support the claim that we achieve competitive performance, on par with the current SOTA or even surpassing it.

{\parskip=3pt
\noindent\textbf{Vehicle segmentation.}
We evaluate our model and multiple SOTA methods for BEV vehicle segmentation in Tab.~\ref{tab:sota}. To ensure a fair comparison, we follow \cite{schramm2024bevcar,chabot2025gaussianbev} and evaluate the task applying vehicle visibility filtering (at least 40\%) and image resolution $(448, 800)$.

We first evaluate against our vision-based baseline, GaussianLSS \cite{lu2025toward}, as well as leading methods including BEVFormer \cite{li2022bevformer}, SimpleBEV \cite{harley2023simplebev}, PointBeV \cite{chambon2024pointbev}, and GaussianBeV \cite{chabot2025gaussianbev}. Our fusion-based approach outperforms them, achieving a +7.0 IoU over the prior SOTA, and demonstrating the added value of radar data in vision-centric approaches. Next, we compare against fusion methods: SimpleBEV \cite{harley2023simplebev}, BEVCar, SimpleBEV++ \cite{schramm2024bevcar}, CRN \cite{kim2023crn}, and BEVGuide \cite{man2023bev}. Note that CRN requires 4 input frames at inference time and, as BEVGuide, does not release code, complicating direct comparison. Our method outperforms SimpleBEV (+1.6 IoU) and performs competitively with the strongest fusion-based approaches, with only a -1.1 IoU gap relative to BEVCar.

}

{\parskip=3pt
\noindent\textbf{Map segmentation.}
For this task, we aim to segment all relevant road elements such as: drivable area, lane boundaries, road and lane dividers, pedestrian crossings, walkways and carpark areas. Following \cite{schramm2024bevcar,chabot2025gaussianbev}, we report metrics for the drivable area and lane boundaries and evaluate the task with image resolution $(448, 800)$ in Tab. \ref{tab:sota-map}.

Our approach surpasses all camera-only baselines in drivable area and lane boundary segmentation, achieving improvements of +0.3 IoU and +2.7 IoU, respectively, over GaussianBeV. When compared to fusion-based methods, our method matches the top-performing approach, BEVCar, in drivable area segmentation, while substantially outperforming it in lane boundary segmentation, with a margin of +4.8 IoU.
}

\begin{figure*}
    \centering
    \includegraphics[width=0.98\textwidth]{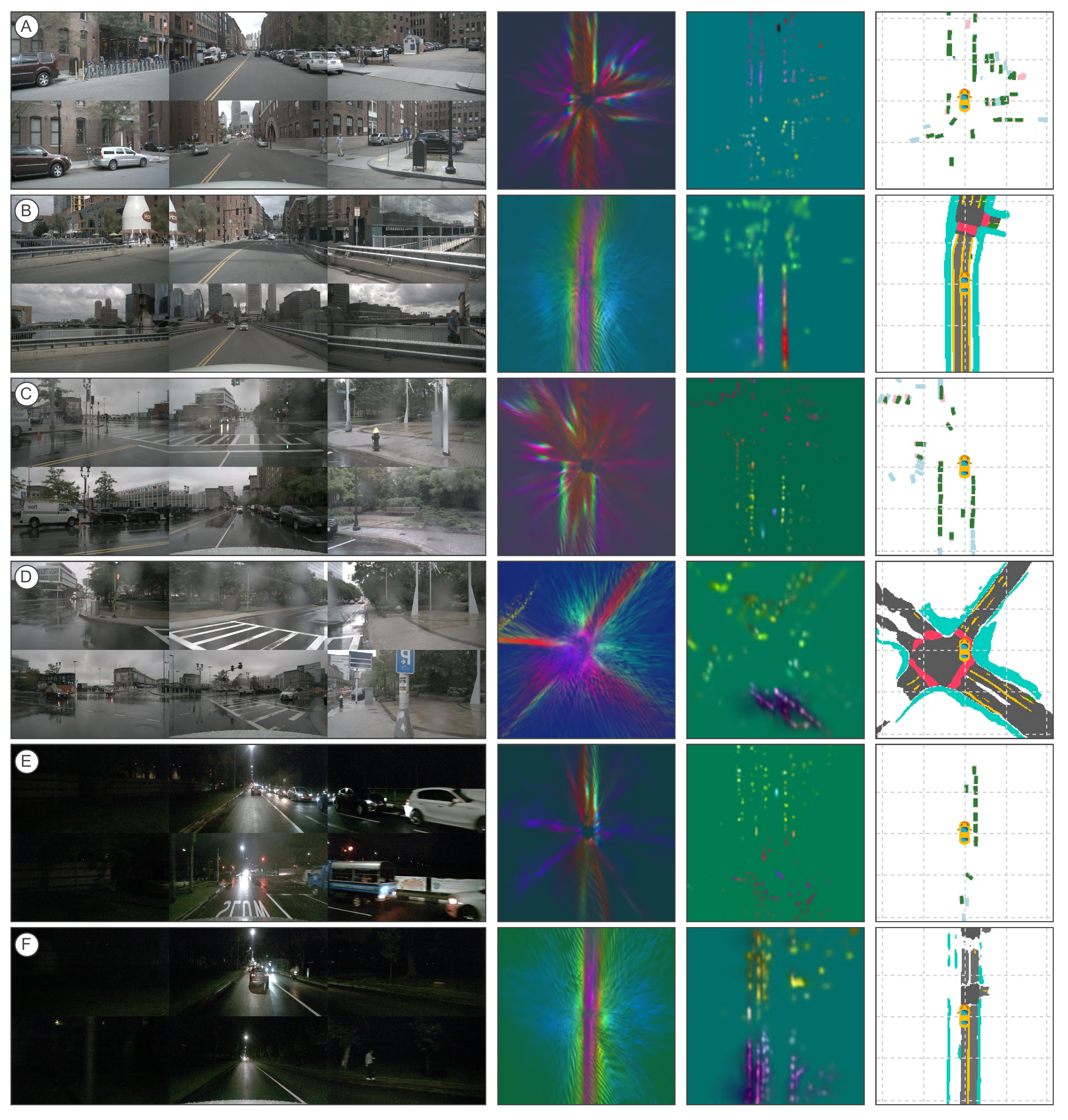}
    \caption{
        \textbf{Qualitative results on the nuScenes validation set.}
        Each row shows, from left to right: multi-view camera images, PCA camera latent
        features, PCA radar latent features, and predictions. For vehicle segmentation, we report an error map where correctness is indicated
        by color: \textcolor{OliveGreen}{correct}, \textcolor{blue!50}{missing}, and
        \textcolor{red!50}{incorrect}. For map segmentation, we report classes by color: \textcolor[RGB]{110, 110, 110}{drivable area}, \textcolor[RGB]{255, 200, 0}{lane and road dividers}, \textcolor[RGB]{255, 61, 99}{pedestrian crossings}, \textcolor[RGB]{0, 207, 191}{walkway} and \textcolor[RGB]{34, 139, 34}{carpark areas}.
    }
    \label{fig:quali}
    \vspace{-0.5cm}
\end{figure*}

\begin{table}[t]
\vspace{2mm}
\centering
\caption{Runtime Analysis} \label{tab:runtime}
\begin{threeparttable}
\begin{tabular}{@{}lccc}
\toprule
\multicolumn{1}{c}{\textbf{Method}}                & Veh. IoU ($\uparrow$) & ms ($\downarrow$) & FPS ($\uparrow$)\\ \midrule
SimpleBEV$^{\dag}$ \cite{harley2023simplebev}     & 55.7             & \textbf{57.6}    & \textbf{17.4}    \\
SimpleBEV++$^{\ddag}$ \cite{schramm2024bevcar}    & 52.7             & 211.3            & 4.7              \\
BEVCar$^{\ddag}$ \cite{schramm2024bevcar}          & \textbf{58.4}    & 245.6            & 4.1              \\ \midrule \rowcolor{gray!15}
\textbf{GaussianCaR}$^{\dag}$ \textit{(vehicle)}   & \underline{57.3} & \underline{75.6} & \underline{13.2} \\ \rowcolor{gray!15}
\textbf{GaussianCaR}$^{\dag}$ \textit{(map)}       & --               & 81.1             & 12.3             \\ \bottomrule
\end{tabular}

\footnotesize{
Inference time of a forward pass measured on an NVIDIA RTX 4090 with image resolution $\dag$: $(448, 800)$, or $\ddag$: $(448, 896)$. Best is marked in \textbf{bold} and second best is \underline{underlined}.
}
\end{threeparttable}
\vspace*{-0.4cm}
\end{table}

\subsection{Ablation Study}

We conduct an incremental ablation study (Tab. \ref{tab:ablation}) to assess our proposed framework, GaussianCaR, which extends the camera-only baseline GaussianLSS \cite{lu2025toward}, evaluating all methods on vehicle segmentation at $(448, 800)$ resolution.

For our baseline, we report an IoU score of $46.1$ with a runtime of $18.6$ Hz. We introduce EfficientViT-L2 as a stronger image backbone to ensure a fair comparison with the current SOTA of camera-radar fusion methods, and the metric offset head. For supervision, we introduce an early guidance loss and a Dice loss component for both segmentation losses, leading to an improvement of $+1.9$ IoU. The combination of these components constitutes our Pixels-to-Gaussians module.

We add a radar branch based on a lightweight PTv3 with 7 accumulated sweeps. Encoding radar positions with a scatter-to-BEV mechanism yields 55.0 IoU, a +7.0 gain over vision-only. Incorporating all radar variables further improves performance by +1.1 IoU. We then introduce Gaussians as a view transformation, allowing features to offset and diffuse locally; together, these components form our Points-to-Gaussians module, reaching 56.9 IoU. Lastly, since we fuse two modalities, we adopted the multi-scale gating fusion and BEV decoder from \cite{montielmarin2025car1} for the previous experiments. We propose to improve this stage by incorporating our CMX-based fusion and DPT-based decoder, yielding +0.4 IoU at 13.2 Hz, confirming SOTA-level accuracy with high efficiency.

To evaluate the effect of accumulated radar sweeps, we ablate the model using 1, 4, and 7 sweeps, obtaining IoU scores of 54.9, 56.2, and 57.3, respectively, which confirms the positive impact of incorporating additional radar sweeps.

\subsection{Runtime Analysis}

To support the claim that our approach is robust and efficient for camera-radar fusion, we report the forward pass runtimes of our proposed \textbf{GaussianCaR} alongside other SOTA methods in Tab.~\ref{tab:runtime}. All measurements were conducted on an NVIDIA RTX 4090 GPU using a batch size of 1 with FP32 precision, and image resolution $(448, 800)$.

A key observation from our method is that, while efficient, the rasterization module scales linearly with the number of Gaussians, making the convergence count a key determinant of runtime. Convergence typically occurs at $\sim$14k Gaussians for the vehicle segmentation task, and $\sim$24k for the map. Consequently, the runtime of our method exhibits slight task-dependent variations. We report a mean runtime of \textbf{75.6 ms} and \textbf{81.1 ms} on vehicle and map segmentation, respectively.

We conduct a fair comparison with two SOTA methods, SimpleBEV++ and BEVCar, both employing similar image backbones. Their inference times are 211.3 and 245.6 ms, respectively. Our method achieves performance on par with BEVCar while preserving the efficiency of SimpleBEV, resulting in a \textbf{3.2× faster runtime} compared to BEVCar.

\subsection{Qualitative Results} \label{subsec:quali}

Fig.~\ref{fig:quali} shows qualitative results on nuScenes. Each row corresponds to a scene, displaying multi-view inputs, PCA-projected camera and radar features, and predictions. Scenes \ref{fig:quali}.a–b represent daytime urban traffic, \ref{fig:quali}.c–d a rainy four-way intersection, and \ref{fig:quali}.e–f nighttime driving.
\section{Conclusion} \label{sec:conclusion}

In this paper, we propose a novel framework for simple, yet robust and efficient camera-radar fusion in perception applications, demonstrating strong performance in both accuracy and inference speed. Our method leverages GS to reframe sensor fusion in latent space as a \textbf{modality} $\xrightarrow[]{}$ \textbf{Gaussians} $\xrightarrow[]{}$ \textbf{BEV} process. We implement and evaluate our approach on BEV segmentation tasks using the nuScenes dataset, achieving performance on par with the SOTA, and even surpassing it in lane divider segmentation. Furthermore, we achieve a 3.2× faster runtime compared to BEVCar, delivering both top-tier performance and fast inference. These promising results also open several avenues for future research, including alternative backbones that exploit the modality-to-Gaussian transformation cycle or novel fusion mechanisms in the Gaussian intermediate space.

\footnotesize
\bibliographystyle{IEEEtran}
\bibliography{references.bib}

\end{document}